\documentclass[letterpaper]{article} 
\usepackage{aaai25}  
\usepackage{times}  
\usepackage{helvet}  
\usepackage{courier}  
\usepackage[hyphens]{url}  
\usepackage{graphicx} 
\usepackage{natbib}  
\usepackage{caption} 
\usepackage{algorithm}
\usepackage{algorithmic}

\usepackage{multicol}
\usepackage{multirow}
\usepackage{amssymb}
\usepackage{xcolor}
\usepackage{arydshln} 
\usepackage{amsmath}
\usepackage{subcaption}
\usepackage{newfloat}
\usepackage{listings}
\DeclareCaptionStyle{ruled}{labelfont=normalfont,labelsep=colon,strut=off} 
\floatstyle{ruled}
\newfloat{listing}{tb}{lst}{}
\floatname{listing}{Listing}
\title{SpikingSSMs: Learning Long Sequences with Sparse and Parallel Spiking State Space Models}
\author {
    Shuaijie Shen\equalcontrib\textsuperscript{,\rm 1,\rm 2},
    Chao Wang\equalcontrib\textsuperscript{,\rm 1,\rm 2},
    Renzhuo Huang\textsuperscript{\rm 1,\rm 2},
    Yan Zhong\textsuperscript{\rm 2,\rm 3}, 
    Qinghai Guo\textsuperscript{\rm 2}, 
    Zhichao Lu\textsuperscript{\rm 4}, \\
    Jianguo Zhang \footnotemark[2]\textsuperscript{,\rm 1,\rm 5},
    Luziwei Leng\thanks{Corresponding author}\textsuperscript{,\rm 2},
}
\affiliations {
    \textsuperscript{\rm 1}
    Department of Computer Science and Engineering,
    Southern University of Science and Technology,
    Shenzhen,
    518055, \\
    
    \textsuperscript{\rm 2}
    ACSLab,
    Huawei Technologies Co., Ltd.,
    Shenzhen, 518129, \\

    \textsuperscript{\rm 3}
    School of Mathematical Sciences,
    Peking University,
    Beijing, 100871, \\

    \textsuperscript{\rm 4}
    Department of Computer Science,
    City University of Hong Kong,
    Hong Kong \\

    \textsuperscript{\rm 5}
    Pengcheng Laboratory, Shenzhen, China \\

    \{shensj2024, wangc2023, huangrz2023\}@mail.sustech.edu.cn, 
    zhongyan@stu.pku.edu.cn,
    guoqinghai@huawei.com,
    zhichao.lu@cityu.edu.hk,
    zhangjg@sustech.edu.cn,
    lengluziwei@huawei.com
}

\begin{document}
\maketitle

\renewcommand{\thefootnote}{\fnsymbol{footnote}}

\begin{abstract}
Known as low energy consumption networks, spiking neural networks (SNNs) have gained a lot of attention within the past decades. While SNNs are increasing competitive with artificial neural networks (ANNs) for vision tasks, they are rarely used for long sequence tasks, despite their intrinsic temporal dynamics. 
In this work, we develop spiking state space models (SpikingSSMs) for long sequence learning by leveraging on the sequence learning abilities of state space models (SSMs). Inspired by dendritic neuron structure, we hierarchically integrate neuronal dynamics with the original SSM block, meanwhile realizing sparse synaptic computation. 
Furthermore, to solve the conflict of event-driven neuronal dynamics with parallel computing, we propose a light-weight surrogate dynamic network which accurately predicts the after-reset membrane potential and compatible to learnable thresholds, enabling orders of acceleration in training speed compared with conventional iterative methods. On the long range arena benchmark task, SpikingSSM achieves competitive performance to state-of-the-art SSMs meanwhile realizing on average 90\% of network sparsity. On language modeling, our network significantly surpasses existing spiking large language models (spikingLLMs) on the WikiText-103 dataset with only a third of the model size, demonstrating its potential as backbone architecture for low computation cost LLMs. 
\end{abstract}

%
\begin{links}
    \link{Code and Supplement}{https://github.com/shenshuaijie/SDN}
\end{links}
\section{Introduction} \label{sec:intro}
Recent years have witnessed the proliferation of real-world time-series datasets in various domains, which often require reasoning over tens of thousands of time steps~\cite{tay2020long}. Therefore, plenty of sequence models have emerged in recent years, which aim to model the long-range dependencies (LRDs) in sequential data to achieve human-level performance across diverse modalities, encompassing text, vision, audio, and video~\cite{gu2021efficiently}. Among these methods, growing attention has been given to Transformer~\cite{vaswani2017attention}, since this architecture has led to remarkable developments in the areas of vision and speech. However, for an input sequence of length $L$, it requires the high-cost computational complexity of $\mathcal{O}(L^2)$ during training and inference in the module of self-attention, which is one of the core contextualizing components in the Transformer model. Although some Transformer variants~\cite{kitaev2020reformer,zaheer2020big,katharopoulos2020transformers,choromanski2020rethinking} are proposed to reduce the compute and memory requirements, their performances on performing long-range reasoning remain considerably suboptimal~\cite{gu2021efficiently}. 

Recurrent neural networks (RNNs) ~\cite{schuster1997bidirectional,sherstinsky2020fundamentals} have emerged early for learning on the variable-length input sequences, which requires only $\mathcal{O}(1)$ operations with respect to the sequence length. However, constrained hidden state space and gradient vanish problem have limited their learning of long sequences. To address this problem, innovative works such as RWKV~\cite{peng2023rwkv} and state space models (SSMs)~\cite{gu2021efficiently,gu2023mamba} are proposed by introducing an appropriate design of hidden states for handling LRDs with both training parallelizability and inference efficiency. 
RNNs owes part of its inspiration to cognitive and neurological computational principles~\cite{lipton2015critical}, which also serve as the foundation for another class of biologically-grounded architectures known as Spiking Neural Networks (SNNs)~\cite{maass1997networks}. With their potential in low-energy computing, SNNs have gained a lot of attention within the past decades. Recently, they have been shown to be as efficient as artificial neural networks (ANNs) for vision tasks \cite{che2022differentiable,zhou2022spikformer,yao2024spike,che2024spatial} under convolution or Transformer architectures. However, despite the intrinsic temporal dynamics, SNNs are rarely used for long sequence tasks. 
Note that SNNs under convolution or Transformer architectures often need a certain simulation time window to improve spike-based representation, causing inference delays compared to their artificial counterparts. This disadvantage can be avoid for SNNs under RNN architecture since they can make use of the inherent temporal dimension for dynamic computing.

In this work, we explore an integration of spiking neurons with SSMs, and develop SpikingSSMs for long sequence learning, combining efficient parallel training and low-energy, spike-based sparse computation.
Several recent works have proposed binary SSM \cite{stan2023learning} or stochastic spiking SSM \cite{bal2024rethinking}. However, they have limited exploration or overlooked the intricate dynamics that characterize biological spiking neurons, leading to incomplete interpretability and performance degradation. To this end, we adopt the widely used Leaky Integrate-and-Fire (LIF) neuron with deterministic reset mechanisms \cite{gerstner2014neuronal}. To reconcile the conflict of its asynchronous event-driven dynamics with parallel computing, we propose a surrogate dynamic network which accelerates training and is dispensable during inference without adding additional parameters to the network. Through an equivalence study we demonstrate the versatility of SDN for approximating parametric LIF neuron models and its potential as general purpose module for parallel computing SNNs. The key contributions of this study are summarized as follows:

\begin{itemize}
    \item We introduce SpikingSSMs for long sequence tasks, which merge the strengths of SSMs in parallel computing and long sequence modeling with sparse computation of SNNs.
    \item To address the challenges posed by event-driven neuronal dynamics in the context of parallel computing, we propose a surrogate dynamic network (SDN) to approximate the dynamics of LIF neurons via a well-designed model, which extremely accelerates the training of SpikingSSMs with only negligible additional computation.
    \item We also highlight the equivalence of SDN for different thresholds and incorporate learnable thresholds into our model architecture, which further improves network performance.
    \item We evaluate our method on sequential and permuted sequential MNIST classification tasks, as well as the Long Range Arena (LRA) benchmark, where our model achieves competitive performance with state-of-the-art SSMs meanwhile with high sparsity. Additionally, in large-scale language modeling task on the WikiText-103 dataset. Our model sets a new record in the field of SNN, demonstrating its scalability.
\end{itemize}
\section{Related Work} \label{sec:related}

\subsection{Long Sequence Modeling}
The essential problem of sequence modeling is compressing context into a certain state. Driven by this problem, sequence models explore trade-offs between efficiency and effectiveness. For example, Attention mechanism ~\cite{vaswani2017attention,dao2022flashattention,dao2023flashattention} does not compress context at all, i.e. it stores the entire context (i.e. the KV cache) during auto-regressive inference, which is effective but inefficient since this causes the slow linear-time inference and quadratic-time training~\cite{sun2023retentive,yang2023gated}. On the other hand, recurrent models compress context into a finite state, resulting in constant-time inference and linear-time training. However, their effectiveness is limited by how well this state has compressed the context and the fixed representation space~\cite{peng2023rwkv,qin2023transnormerllm}. SSMs have emerged as compelling frameworks for sequence modeling. HiPPO~\cite{gu2020hippo} revolutionized this field by compressing long inputs into dynamic, polynomial-based representations using orthogonal polynomials. S4~\cite{gu2021efficiently} further evolved this approach by introducing a low-rank correction, enabling stable diagonalization and simplifying operations with Cauchy kernels. A series of later works 
have further improved efficiency of the model using advanced techniques such as parallel scan \cite{smith2022simplified}, Fast Fourier Transform (FFT) \cite{fu2022hungry, duhamel1990fast} and gating mechanism \cite{mehta2022long}.
A very recent work, Mamba~\cite{gu2023mamba} focuses on enhancing the selectivity of the state representation, balancing efficiency and effectiveness without compromising contextual information. Aided with hardware-optimized algorithms the model demonstrated strong performance on temporal tasks up to million-length sequences such as language modeling.

\subsection{SNNs for Sequence Modeling}
With the improvement of SG training methods, SNNs adopting conventional RNN architectures have been applied to sequence classification tasks and achieved high accuracy \cite{bellec2018long,yin2021accurate,yin2023accurate}. However, limited by the architecture and serial processing, pure RNN-based SNNs are rarely applied to long sequence learning. To this end, enabling efficient parallel computing of SNN is critical. PSN \cite{fang2024parallel} achieved it by removing the reset of spiking neuron, however with the cost of increased firing rate and insufficiency in network sparsity. PSU  \cite{li2024efficient} proposed parallel spiking units which decoupled the integration-spiking-resetting process by introducing a probabilistic reset mechanism and effectively improved network sparsity. However, its learnable parameter is quadratic to the sequence length which impeded the scalability of the method. Leveraging on the Legendre Memory Units (LMU) for sequence modeling \cite{voelker2019legendre}, SpikingLMUFormer \cite{liu2024lmuformer} augmented the LMU with convolutional layers and spiking activation, surpassing transformers in long sequence modeling. The recent progress of SSMs has also inspired works developing their spiking versions. 
\citeauthor{du2024spiking} proposed SpikeS4 by simply stacking LIF neurons on S4 layers and applied for speech tasks. Binary S4D \cite{stan2023learning} constructed binary SSM by directly applying spiking activation function on the summation of hidden states, which maintains parallel training but ignores neuronal dynamics and sparsity. A recent work \cite{bal2024rethinking} proposed S6-based SNN which improved network sparsity by implementing a stochastic spiking neuron for SSM, however the model exhibited significant accuracy drop compared to the original model, partially attributed to the stochastic noise in gradients. In this work, we adopt widely used deterministic reset dynamics for spiking neurons, and develop solutions to solve the conflict of their asynchronous event-driven feature with parallel computing.

\subsection{SNNs for Language Modeling}
Motivated by the potential of constructing low-energy large language models, several recent works have explored combining SNNs with language models. SpikeGPT \cite{zhu2023spikegpt} adopted spike activation for the output of RWKV \cite{peng2023rwkv} blocks and applied to large scale language modeling tasks.
SpikeBERT \cite{lv2023spikebert} built upon Spikformer \cite{zhou2022spikformer} and distilled knowledge from the original BERT \cite{devlin2018bert}. In this work, we develop large scale SNNs based on SSM architectures for language modeling.

\section{Method} \label{sec:method}

\subsection{Preliminaries}
\subsubsection{LIF Neuron}
The LIF neuron is a simplified version of biological neuron models \cite{gerstner2014neuronal}, which captures the "leaky-integrate-fire-reset" process and is widely used in SNNs for machine learning as it balances tractability and temporal dynamics. With $t$ denoting the time step, the LIF neuron is formulated by following equations:
\begin{equation}
\label{eq:lif-mem}
    u'_{t} = \tau u_{t - 1} + I_t
\end{equation}
\begin{equation}
\label{eq:spike}
    s_t = H(u'_t - v_{\mathrm{th}})
\end{equation}
\begin{equation}
\label{eq:soft-reset}
    \mathrm{Soft \, reset}: \, u_t = u'_t - s_t v_{\mathrm{th}}
\end{equation}
\begin{equation}
\label{eq:hard-reset}
    \mathrm{Hard \, reset}: \, u_t = u'_t(1 - s_t) + s_tu_r
\end{equation}
where input currents $I$ are linearly integrated into the leaky membrane potential $u$ of the neuron, and then a spike $s$ is determined to be fired if the current $u$ surpasses a threshold $v_{\mathrm{th}}$, with $H$ denoting the Heaviside function. At last, the membrane potential is reset according to the soft reset mechanism (Eq. \ref{eq:soft-reset}) or the hard reset mechanism (Eq. \ref{eq:hard-reset}).
The hard and soft reset mechanisms embody different neuronal memory strategies, where the hard reset forget the history after spiking and reset to a reset potential $u_r$ (we set it to 0 in this work), while the soft reset still keeps all the history subtracted by a reset after spiking. In order to realistically mimic biological neurons, the hard reset mechanism is most commonly used in spiking networks.

\subsubsection{Surrogate Gradient Training of SNN}
Since the spikes are considered identical, the spiking activation function $H$ is defined as a Heaviside function which is non-differentiable at $x = 0$ and has a derivative value of 0 elsewhere. Therefore, surrogate gradient (SG) methods \cite{sg1,bellec2018long} are proposed to solve this issue. The surrogate gradient function is defined as a soft relaxed function that approximates the original discontinuous gradient of the spiking activation function. Typical SG functions are usually differentiable everywhere and have a nonzero derivative value near the threshold, such as rectangular \cite{zheng2021going} and triangular \cite{bellec2018long} functions, etc.

\subsubsection{State Space Model}
SSMs are broadly used in many scientific disciplines, which map a 1-dimensional signal $x$ to an N-dimensional latent signal $h$ and project it to a 1-dimensional output signal $y$.
For a discrete input sequence $x_{1:L}$, through certain discretization rule \cite{gu2022parameterization} the SSMs can be defined by:
\begin{equation}
    \label{eq:ssm-x}
    h_t = \bar A h_{t-1} + \bar B x_t
\end{equation}
\begin{equation}
    \label{eq:ssm-y}
    y_t = Ch_t
\end{equation}
with subscript $t$ denoting the time step. The parameters are the state matrix 
$\bar A \in \mathbb{R}^{N \times N}$ and other matrices $\bar B \in \mathbb{R}^{N \times 1}$, $C \in \mathbb{R}^{1 \times N}$. Theoretically, $\bar A$ can be diagonalized for efficient computation \cite{gupta2022diagonal}. 
Within a layer of the network, the input is always multidimensional rather than 1-dimension, therefore, an SSM layer handles multiple features by multiple independent SSMs (with different parameters).
In parallel computing, the SSM can be expressed as the convolution between convolution kernels and input signals, with the initial condition $y_0 = 0$:
\begin{equation}
    \label{eq:ssm-conv}
    y_t = \sum_{k = 1}^{t}C \bar A^{t - k} \bar B x_k
\end{equation}
In practice, this computation can be further accelerated by FFT with time complexity $\mathcal{O}(L\mathrm{log}(L))$ \cite{gupta2022diagonal}].

\subsection{Spiking S4 Block}
\begin{figure*}[htp!]
  \centering
  \includegraphics[width=1.\linewidth]{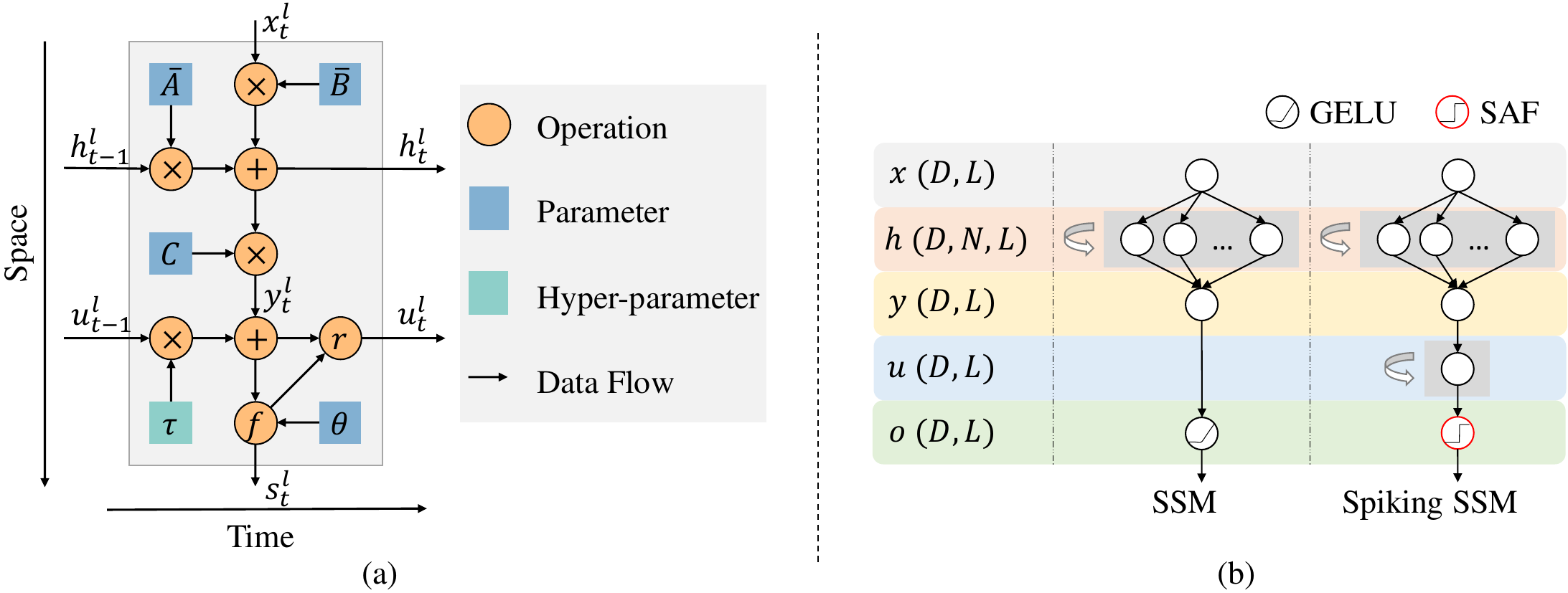}
\caption{Architecture of SpikingSSM. (a) Forward computation graph of SpikingSSM in one layer. Operation $r$ denotes the reset mechanism. The learnable parameter $\theta$ denotes parameters that influence the spiking function $f$, such as the threshold. (b) Comparison of different SSMs. The original SSM outputs float point number. SpikingSSM replaces the non-linear function of original SSM with an LIF neuron, adding neuronal dynamics on a higher hierarchy. SAF denotes the spiking activation function. The left panel denotes the computation stage of different variables and their corresponding dimensions, with $D, N, L$ denoting the model dimension, the hidden dimension of SSM and the sequence length, respectively.}
\label{fig:spiking-s4}
\end{figure*}
It has been shown that the diagonal version of SSM \cite{gupta2022diagonal} maintains performance while simplifying the model. Therefore, we choose the latest S4D model \cite{gu2022parameterization} as the backbone to verify our method. The output $y$ of the state space block is now activated by an LIF neuron, i.e. the $y_t=Ch_t$ is treated as the input current of the neuron:
\begin{equation}
\label{eq:ssm-mem}
    u'_{t} = \tau u_{t - 1} + y_t
\end{equation}
\begin{equation}
\label{eq:ssm-spike}
    s_t = H(u'_t - v_{\mathrm{th}})
\end{equation}
The spiking output is then feed into the FC layer of the next spiking S4 block, which undergoes addition operation with the weight matrix, realizing low-energy, sparse synaptic computation. The threshold largely controls the spiking rate of the neuron, inspired by previous works \cite{rathi2021diet}, we set it as a learnable parameter during training to optimize network performance. A comparison of different s4 blocks and the spiking s4 block is shown in Fig. \ref{fig:spiking-s4}. Interestingly, from a neurobiological perspective, the structure of the spiking S4 block resembles a multi-time scale dendritic neuron \cite{london2005dendritic,zheng2024temporal}, with $h$ representing dendrites and $y$ representing the soma which receives collective input from dendrites, both characterized by self-recurrent temporal dynamics.

\subsection{Surrogate Dynamic Network}
Since $y$ can be calculated in parallel, given an input sequence $y_{1:T}$ to the spiking neuron, under hard reset, $u_t$ with $t \in [1,T]$ can be formulated as:
\begin{equation}
\label{eq:ssm-mem-hardreset}
    u_t=\sum_{i=1}^t \bigg[ \prod_{j=i}^{t-1}(1-s_j) \cdot \tau^{t-i} \cdot y_i \bigg]
\end{equation}
It can be seen that the membrane potential is determined by the past spiking history of the neuron which can not be computed in parallel, thus SNNs always adopt the form of iterative computing. The nonlineariy of spiking activation, especially the event-driven reset mechanism prevents parallel computing of SNNs, which makes them not practical for efficient training on modern hardware, especially for long sequence tasks. 
The neural networks, however, are designed for handling the mapping between inputs and outputs, and can be parallelized on modern hardware. 
Since spiking neurons with fixed parameters must produce the same outputs with the same inputs, the 'neuron' can be considered as a black-box that maps the input to spike sequence, which is exactly what neural networks are good at. Therefore, we propose using a pre-trained neural network, dubbed as Surrogate Dynamic Network, to predict the spike train in parallel. Specifically, we train a network $f$, which learns the neuronal dynamics that maps input to output spike trains. For example, a neural network to predict spike train based on all time-step input can be expressed as:
\begin{equation}
\label{eq:pred-s}
    s_{1:T} = f(I_{1:T})
\end{equation}
where $I_{1:T}$ is the input current from time-step $1$ to $T$, and $s_{1:T}$ is the corresponding spike train predicted by the network $f$.

\begin{figure}[t!]
    \centering
    \includegraphics[width=1.0\linewidth]{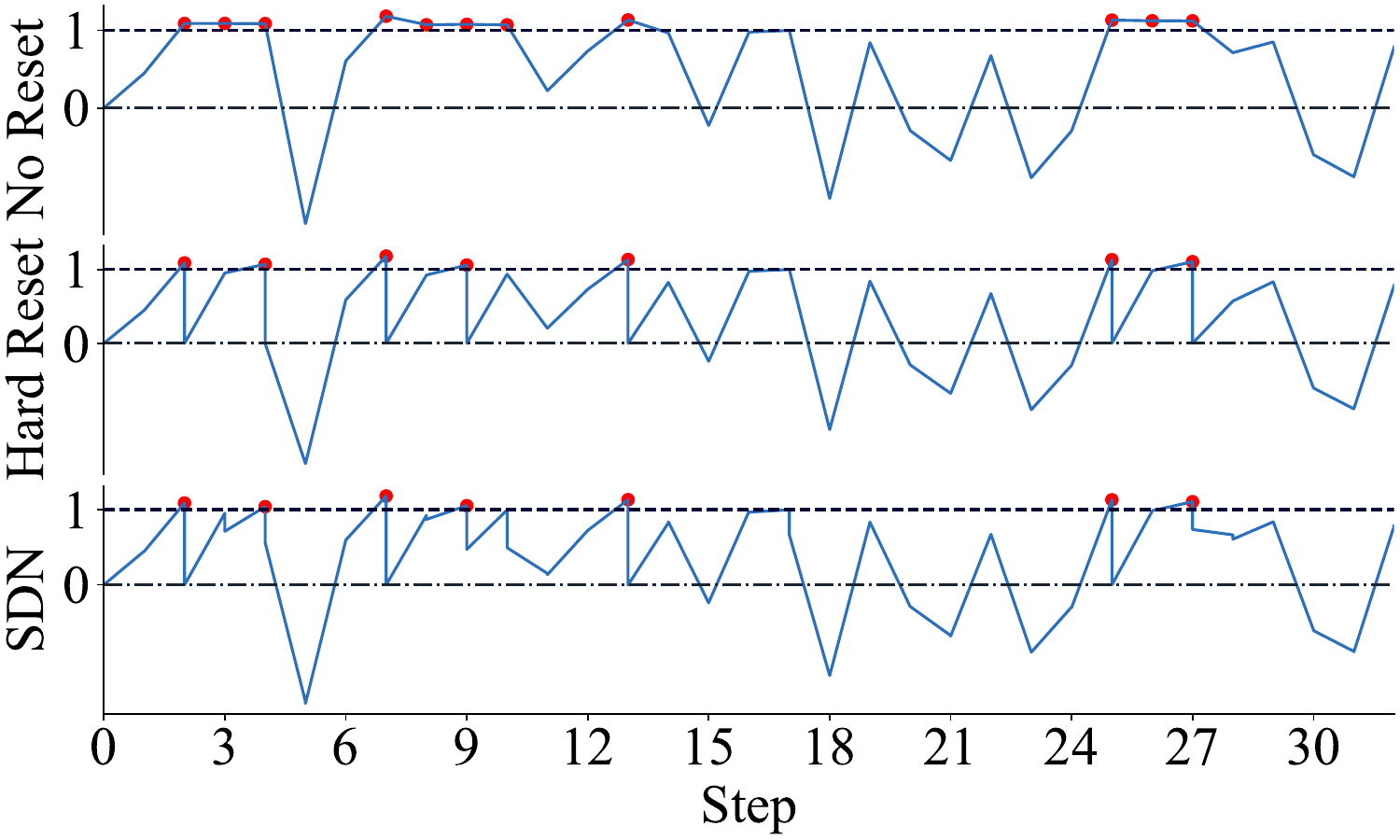}
    \caption{Comparison of membrane potential samples produced by different methods under the same input. The membrane potential predicted by the SDN (bottom) accurately approximates the ground truth produced by the spiking neuron (middle). Without reset the membrane potential significantly produces more spikes (top). The two black dashed lines denote the reset potential and the spiking threshold which are set to 0 and 1, respectively. Red points denote moments when spikes are generated, i.e. the membrane potential surpasses the threshold. Note that for the spiking neuron, the membrane potential is reset to 0 immediately once surpasses the threshold. }
    \label{fig:mem-pot}
\end{figure}

Meanwhile, for the sake of efficiency, the network should be very small such that the forward inference can be done with low computation cost. As demonstrated in the experiment, a 3-layer network with 1-D convolution is sufficient to learn the neuron dynamics and accurately predict the spike, as shown in Fig. \ref{fig:mem-pot} (more details are presented in the experiment section).
To further accelerate training and simplify the computational graph, we switch the trained SDN to inference mode without backpropagation during training of the task network, using its predicted spike train and the input to compute the membrane potential as equation \ref{eq:ssm-mem-hardreset}. Finally, the spike is determined by the membrane potential as the output of the spiking S4 block. 
During test mode, the SDN can be either kept for parallel inference with linear time complexity with respect to the sequence length, or removed for real-time iterative inference with time complexity $\mathcal{O}(1)$, without adding additional parameters to the network, i.e. spiking neurons switch to the original reset mechanism.
Note that in order to reduce the complexity of computational graph, the SDN can also be trained to predict the membrane potential after leaking, i.e., the '$\tau u_{t-1}$' term in equation \ref{eq:ssm-mem}. In this case, the computational graph has a similar form as in spatial learning through time (SLTT) \cite{SLTT}, which has been demonstrated effective and more efficient than the traditional backpropagation through time (BPTT) for the training of SNN. More details of the derivation are provided in Appendix \ref{appdix:graph}.

\subsection{Learnable Threshold}
The threshold determines the moment of spike generation and largely modulates the spiking rate of SNN. Previous works \cite{rathi2021diet,wang2022ltmd} have shown that optimizing the threshold during the training of SNN can improve network performance. Can SDN approximate neuron dynamics with different threshold during the training of SpikingSSM? We demonstrate that this is feasible through an equivalence study. 
First, we identify some important properties of the threshold, given that both the initial membrane potential and the reset potential are $0$.

\textbf{Property 1. The ratio of inputs and threshold determines the dynamic process of the neuron.}

In other words, if we scale the threshold and inputs with the same factor, the spike train will remain unchanged. Formally, if $f$ represents the dynamic process of the neuron, we have:
\begin{equation}
\label{eq:dynamic-process}
    s_{1:T} = f(I_{1:T}; v_{\mathrm{th}}) = f(\alpha I_{1:T}; \alpha v_{\mathrm{th}})
\end{equation}

\textbf{Property 2. The threshold scales the distribution of the input.}

For neurons with different threshold, the threshold functions as a scaling factor, therefore we can get a general SDN that only acts as a 'neuron' with $v_{\mathrm{th}} = 1.0$ by fed scaled inputs $\frac{I_{1:T}}{v_{\mathrm{th}}}$ (More details are provided in Appendix \ref{appendix:vth}).
Therefore, based on these properties, we can incorporate a trainable threshold for SDN by learning a scaling factor for the input.

\section{Experiments} \label{sec:experiment}
In this section, we first introduce the architecture design, training and evaluation of SDN. In addition, we benchmark SpikingSSM assisted by SDN against traditional iterative training approaches on training speed. Next, we validate SpikingSSM on three benchmarks tasks of different scales, including classification on the sequential MNIST dataset and its permuted variant, long sequence modeling on the LRA dataset, and language modeling  on the WikiText-103 dataset. Finally, we perform ablation studies of our method and analyze the computation cost of the model.

\subsection{Training and Evaluation of SDN}\label{sec:exp-nn}
\subsubsection{Dataset}
The training dataset for SDN contains the input currents and their corresponding target membrane potentials. The inputs $\in R^L$ are sampled from normal distribution $N(0, 1)$, where $L = 1024$ is the sequence length.
The ground truth membrane potentials are provided by iterative LIF neuron with hard reset. The number of training and testing samples are $10^5$ and $10^4$, respectively.

\subsubsection{Architecture of SDN}
The SDN is a 4-layer CNN constructed by 1-D convolutions and 1-D batch normalizations, denoted by "C8k1s1p0g1-C8k8s1p8g8-Trunc-BN-relu+C8k1s1p0g1-BN+relu-C1k1s1p0g1", where "C", "k", "s", "p" and "g" denote output channel, kernel size, stride, padding and group, with the numbers following them indicating the value. The term "Trunc" signifies truncating the input to maintain a constant length, and the two "+" symbols denote the start and end of a residual connection.

In this case, the total number of parameters in SDN is less than 200, which is minor compared with the backbone network. More details about the architecture are provided in Appendix \ref{appendix:arch}.

\subsubsection{Fitting Ability}
\begin{figure}[h]
    \centering
    \includegraphics[width=1.0\linewidth]{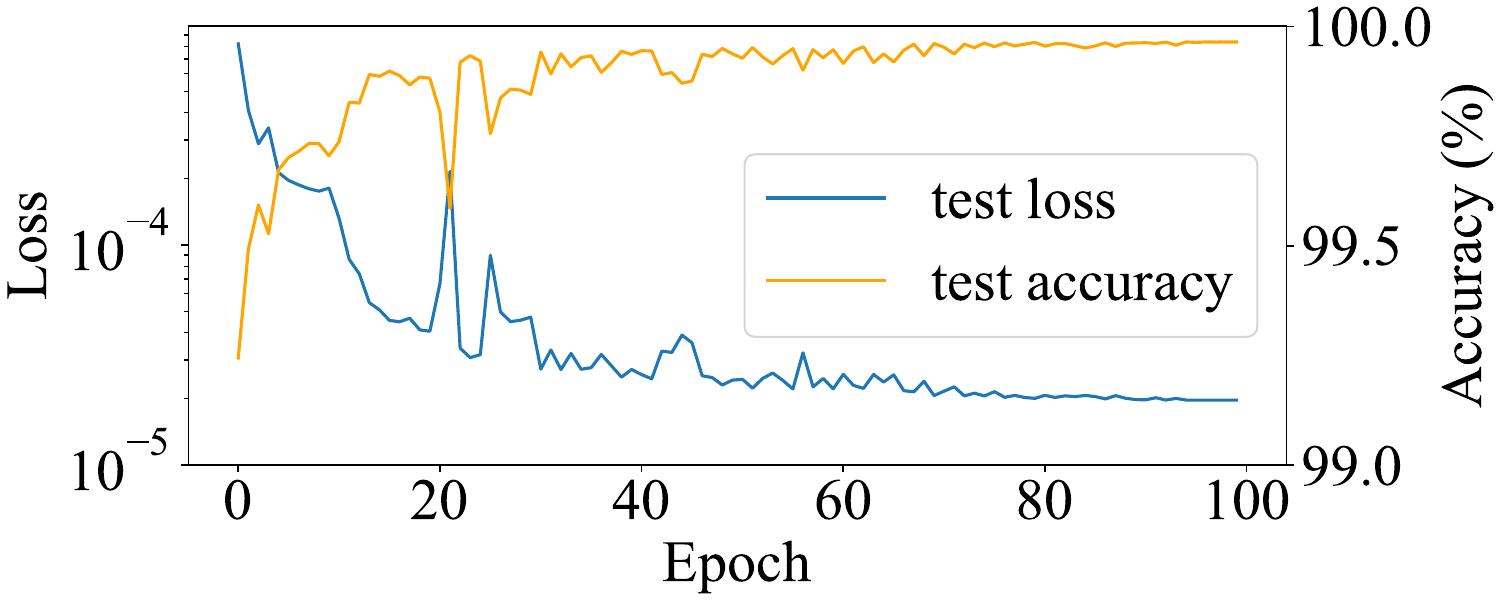}
    \caption{Training of SDN. The MSE loss and spiking accuracy on the test set are plotted here. Note that SDN already achieves sufficiently high accuracy after the first training epoch.}
    \label{fig:loss-acc}
\end{figure}
We train SDN on the generated dataset with mean square error (MSE) as the loss function for 100 epochs. For testing, we further evaluate the spiking accuracy from the predicted membrane potential by comparing with the spikes generated from the ground truth membrane potential. Fig. \ref{fig:loss-acc} shows that the loss of SDN converges and the model gradually attains high spiking accuracy. We also present samples of membrane potentials predicted by SDN for better illustration. As shown in
Fig. \ref{fig:mem-pot}, the membrane potential predicted by SDN closely approximates the ground truth. Without reset, the membrane potential significantly produces more spikes. Note that in some cases the network could mistakenly reset the membrane potential in a minor degree. This occasionally happens when the membrane potential is very close to the threshold, e.g. at the 3rd step. The result proves that SDN can accurately model the membrane potential dynamics of the LIF neuron. Although there is still minor difference between the predicted value and the ground truth, it has negligible impact on the final trained network performance, as demonstrated in the ablation study.

\subsubsection{Comparison on Training Speed}
\begin{table}[h]
\centering
\begin{tabular}{lrrrr}
\hline 
\multirow{2}{*}{\textbf{Method}} &  \multicolumn{4}{c}{\textbf{Speed (ms)}} \\
& \textbf{$L = 1K$}   & \textbf{$L = 2K$} & \textbf{$L = 4K$} & \textbf{$L = 8K$}\\
\hline
BPTT   & $1370$  & $2900$  & $8040$  & $25600$\\
SLTT   & $1210$  & $2720$  & $7740$  & $25600$\\
Ours   & $183$  & $196$   & $200$    & $253$\\
\hline
Ratio & $7.5 \times$ & $15.0 \times$ & $40.2 \times$ & $101.2 \times$ \\
\hline
\end{tabular}
\caption{Comparison on training speed of different methods. The input has a batch size of $64$. Training with SDN achieves significant acceleration, the speed up ratio amplifies with increasing sequence length.}.
\label{tb:benchmark}
\end{table}

We compare the training speed of SpikingSSM assisted by SDN with traditional training methods based on iterative LIF neurons, including BPTT and the more recent SLTT with optimized computational graph. The inputs are 1-D sequences with varying lengths of $L=1K, 2K, 4K$, $8K$ with batch size of $64$. The time measurement is done on a single GPU. As shown in Table \ref{tb:benchmark}, the speed up ratio using SDN amplifies with increasing sequence length, achieving two orders of acceleration at $8K$. Therefore, SDN extremely accelerates the training of SpikingSSM, especially for long sequences. More details about the breakdown of latency are provided in Appendix \ref{appendix:more}.

\subsection{Long Sequence Tasks with SpikingSSM} \label{sec:sssm}
\subsubsection{Sequential MNIST}
The MNIST dataset \cite{lecun1998mnist} comprises 70,000 grayscale images of handwritten digits (0-9), divided as 60,000 training and 10,000 testing images each with a size of 28$\times$28 pixels. The sequential MNIST (sMNIST) dataset \cite{le2015simple} is created by flattening the original 2-dimensional images into sequences of 784 elements. And the permuted sequential MNIST (psMNIST) variant \cite{le2015simple} applied a fixed permutation to the pixels, thereby distorting the temporal structure within the sequence. 
As shown in table \ref{MNIST CIFAR}, SpikingSSM demonstrates competitive performance with other works on both sMNIST and psMNIST datasets.

\begin{table}[htbp]
\centering
\begin{tabular}{l c c c}
\hline
\textbf{Model} & \textbf{SNN} & \textbf{sMNIST}  & \textbf{psMNIST}  \\ \hline
LMUformer         &No    &---          &98.55           \\  
S4                &No     &99.63        &98.70           \\ \hline
SpikingLMUformer        &Yes    &---          &97.92           \\  
Binary-S4D        &Yes    &99.4         &---       \\ 
S6-based SNN      &Yes     &---          &98.4          \\ 
SpikingSSM  &Yes     &99.6             &98.4        \\
\hline
\end{tabular}
\caption{Performance comparison of SpikingSSM and other works on sMNIST and psMNIST datasets.}
\label{MNIST CIFAR}
\end{table}

\subsubsection{LRA}
\begin{table*}[h]
\centering
\begin{tabular}{l c c c c c c c c c}
\hline
\textbf{Model} & \textbf{SNN} & \textbf{LISTOPS}  & \textbf{TEXT} & \textbf{RETRIEVAL} 
&\textbf{IMAGE} & \textbf{PATHFINDER}  & \textbf{Path-X} & \textbf{AVG} \\ \hline
Transformer &No &36.37 &64.27 &57.46 &42.44  &71.40  &---  &53.66 \\ 
LMUFormer &No&34.43 &68.27 &78.65 &54.16  &69.90  &---   &59.24 \\ 
S4D-Lin &No  & \textbf{60.52}  & \textbf{86.97} &  \textbf{90.96} &  \textbf{87.93}  &  \textbf{93.96} &  \textbf{92.80$^\ast$}   &  \textbf{85.52}\\  \hline
Spiking LMUFormer   &Yes &37.30 &65.80 &79.76 &55.65  &72.68  &---  &60.20\\ 
Binary S4D &Yes  &54.80  & \textbf{82.50} &85.03 &82.00  &82.60 &61.20   & 74.69 \\ 
S6-based SNN &Yes  &55.70  &77.62 &88.48 &80.10  & 83.41 &---  & 72.55\\  \hdashline
    SpikingSSM-VF
    & \multirow{2}{*}{Yes}  
    &59.93
    &82.35
    &88.20
    &86.81 
    &\textbf{93.68}
    &94.80
    &84.30 \\
    
    \textcolor{blue}{\small(spiking rate)}
    &
    &\textcolor{blue}{\small(0.13)}
    &\textcolor{blue}{\small(0.10)}
    &\textcolor{blue}{\small(0.06)}
    &\textcolor{blue}{\small(0.22)}
    &\textcolor{blue}{\small(0.07)}
    &\textcolor{blue}{\small(0.10)}
    &\textcolor{blue}{\small(0.11)}\\ \hdashline
    
    SpikingSSM-VT
    & \multirow{2}{*}{Yes}  
    &\textbf{60.23}
    &80.41
    &\textbf{88.77}
    &\textbf{88.21} 
    &93.51
    &\textbf{94.82}
    & \textbf{84.33} \\
    \textcolor{blue}{\small(spiking rate)}
    & 
    &\textcolor{blue}{\small(0.14)}
    &\textcolor{blue}{\small(0.06)}
    &\textcolor{blue}{\small(0.06)}
    &\textcolor{blue}{\small(0.15)}
    &\textcolor{blue}{\small(0.08)}
    &\textcolor{blue}{\small(0.10)}
    &\textcolor{blue}{\small(0.10)} \\ \hline
\end{tabular}
\caption{Performance comparison of SpikingSSM and previous works on the LRA dataset. $^\ast$Since the original S4D-Lin failed in the Path-X task, we instead present the result of another close variant S4D-Inv. -VF and -VT denote fixed and trainable threshold, respectively. Furthermore, we take the 50\% accuracy for the absence of Path-X accuracy as did in the work of S4D, then compute the overall average metrics across all tasks as AVG. The spiking rate for each task have also been calculated, which is indicated by blue font.}
\label{tb:LRA}
\end{table*}
The LRA benchmark \cite{tay2021long} is proposed for the purpose of benchmarking sequence models under the long-context scenario. LRA comprises six tasks featuring sequences that range from $1K$-$16K$ steps, spanning various modalities such as visual data, mathematics experssions, and text. These tasks are designed to assess model abilities in long-context understanding including text classification, document retrieval, image classification, pathfinder, and listops. Table \ref{tb:LRA} compares SpikingSSM against both non-spiking and spiking networks with transformer or SSM architectures. The SpikingSSM adopts a similar architecture as the original S4D model with parameter initialization as in S4D-Lin \cite{gu2022parameterization}. While maintaining a level of accuracy comparable to that of the original model, the SpikingSSM achieves almost 90\% of average network sparsity. Our model also demonstrates a significant performance improvement over previous SNN sequence models. 
Notably, SpikingSSM successfully tackles the Path-X task, a highly challenging problem that requires reasoning over long-range dependencies within sequences of length 128 $\times$ 128, totaling 16,384 steps. Our SpikingSSM with a trainable threshold shows better overall performance and sparsity compared to a fixed threshold. Through further analysis, we find that the trainable threshold 
facilitates a bimodal distribution of the membrane potential, which reduces quantization error of spiking and improves information transmission of the SNN, consistent with previous findings \cite{guo2022recdis} (details are provided in Appendix \ref{appendix:mem-dis}).

\subsubsection{WikiText-103} 
The WikiText-103 dataset is a comprehensive collection of text from Wikipedia articles rated as Good or Featured, consisting of over 100 million tokens spanning a diverse range of topics and domains. 
We adopted the commonly used perplexity as the metric.
Due to its composition of full articles, this dataset is particularly well-suited for models designed to capture long-term dependencies, making it a critical benchmark for word-level language modeling. In our experiments, we adopted a more parameter-efficient setup compared to the S4 model (details are provided in Appendix \ref{appendix:setting}). Despite utilizing significantly fewer parameters, the SpikingSSM, not only outperforms the pre-trained SpikeGPT, but also substantially narrows the performance gap with ANN networks.

\label{table wt-103}
\begin{table}[h]
\centering
\begin{tabular}{l c c r}
\hline
\textbf{Model} & \textbf{SNN}  & \textbf{PPL}   & \textbf{Parameters}\\ \hline
Transformer     &  No    &20.51   & 231M   \\ 
S4               &   No   &20.95      & 249M \\ \hline
SpikeGPT   & Yes &39.75    &213M    \\ 
SpikingSSM  &  Yes &  33.94     & 75M \\ \hline
\end{tabular}
\caption{Performance comparison of SpikingSSMs with previous works on WikiText-103 dataset.}
\label{Wikitext}
\end{table}

\subsection{Ablation Study}
To verify the important roles of SDN, we conduct an ablation study on whether replacing LIF neurons with SDN in SpikingSSM during training causes performance degradation. In addition, as a pre-trained network, SDN has learned to model the dynamics of LIF neurons, and this bias restricts SDN to act as the LIF neuron, but does this bias really help the performance of SpikingSSM? 
We build three models with identical architecture and same hyperparameters, expect the spiking activation function. 'LIF' uses the iterative LIF neuron, 'SDN-S' uses SDN that is trained from scratch with the SpikingSSM end-to-end, and 'SDN' uses the fixed pre-trained SDN. We train these three models on sCIFAR10 dataset (the IMAGE subset in LRA). Table \ref{tb:ablation-SDN} shows the results of these three models. The 'SDN' model achieves comparable performance to the iterative LIF neuron and greatly accelerates training. The 'SDN-S' model fail in achieving comparable performance to 'SDN', demonstrating that the bias of restricting SDN to act as the LIF neuron is beneficial.

\begin{table}[h]
\centering
\begin{tabular}{lccc}
\hline
Model & Accuracy (\%)  & Spiking Rate (\%) & Speed (ms)\\
\hline
LIF   & 85.45   &  12.08    & 1480 \\
SDN  & 85.57   &  11.92 & 230  \\
SDN-S   & 81.52   &  18.30 & 285 \\
\hline
\end{tabular}
\caption{Performance comparison on the sCIFAR10 dataset.}
\label{tb:ablation-SDN}
\end{table}

\subsection{Computation Cost}
\begin{figure}[t!]
    \centering
    \includegraphics[width=1.0\linewidth]{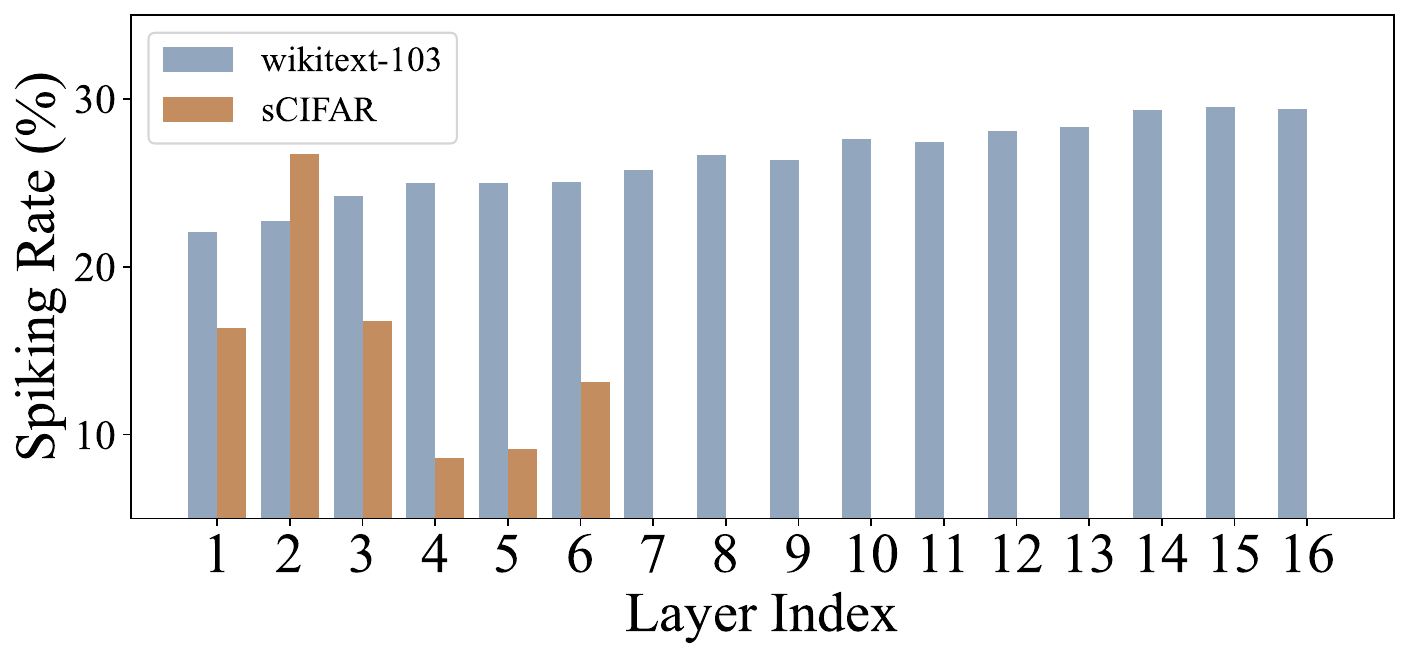}
    \caption{Spiking rate across all layers of SpikingSSMs on the sCIFAR10 and the WikiText-103 datasets.}
    \label{fig:SR}
\end{figure}
Spiking networks are considered low energy cost because the activation of spiking neurons are binary value, and the multiplication between binary activation value and float number weight can be done via only addition operation in some neuromorphic chips, e.g., Speck\cite{Speck}. 
Therefore, the major operation synaptic Accumulation (AC) in SNN has lower energy cost compared to the major operation Multiply-and-Accumulate (MAC) in ANN. Although the hardware implementation and the dynamics of spiking neurons are ignored, a theoretical energy consumption analysis gives an estimation of the efficiency of SNN. Refer to previous works\cite{Speck,FMI}, we assume the energy cost of MAC $E_{\mathrm{MAC}} = 4.6 pJ$ and AC $E_{\mathrm{AC}} = 0.9 pJ$\cite{Energy}.

We define spiking rate as the ratio of the number of spikes to the total time steps of a neuron; the mean spiking rate of the whole network is the mean of spiking rate of all neurons in network.
We denote spiking rate as the mean spiking rate.
Fig. \ref{fig:SR} shows the spiking rate of each layers.
Note that the parameters and computation are mainly from the feature-mix layers, we list the MAC, AC and energy cost in these layers.
For the WikiText-103 dataset with sample length $L = 8192$, our model has 16 layers, in which a linear layer projecting the spikes from $d = 1024$ to $d = 2048$. If all projections are done via multiplication between float numbers, it contains $275.2 G$ MAC, and requires about $1.265 J$. However, the inputs of these layers are binary numbers in our model, and the average spiking rate is less than 30\%. According to the spiking rates in Fig. \ref{fig:SR}, our model contains $72.66G$ AC, and requires about $65.40 mJ$.

\section{Conclusion}
In conclusion, by hierarchically integrating the LIF neuronal dynamics with SSMs, we propose the SpikingSSM which shows competitive performance in long-sequence learning with efficient sparse computation of SNNs. 
For the efficient training of SNN with iterative LIF neurons, we propose a surrogate dynamic network to approximate the dynamics of LIF neurons with parallel computing, which extremely accelerates the training of SpikingSSMs. The SDN is switched to inference mode in training the task spiking networks with only negligible additional computation. We also demonstrate the versatility
of SDN for approximating parametric LIF neuron models and its potential as general purpose module for parallel computing SNNs. 
The application of SpikingSSMs to various benchmark tasks, including the LRA and WikiText-103, not only showcase their competitive performance against previous works but also emphasizes their advantages in sparsity and low-energy requirements. This study contributes to the broader applicability of spiking neural networks, especially in fields requiring efficient processing of long sequence data.

\appendix

\section{Computational Graphs of BPTT, SLTT and SDN} \label{appdix:graph}
\begin{figure*}[h!]
    \centering
    \includegraphics[width=0.75\linewidth]{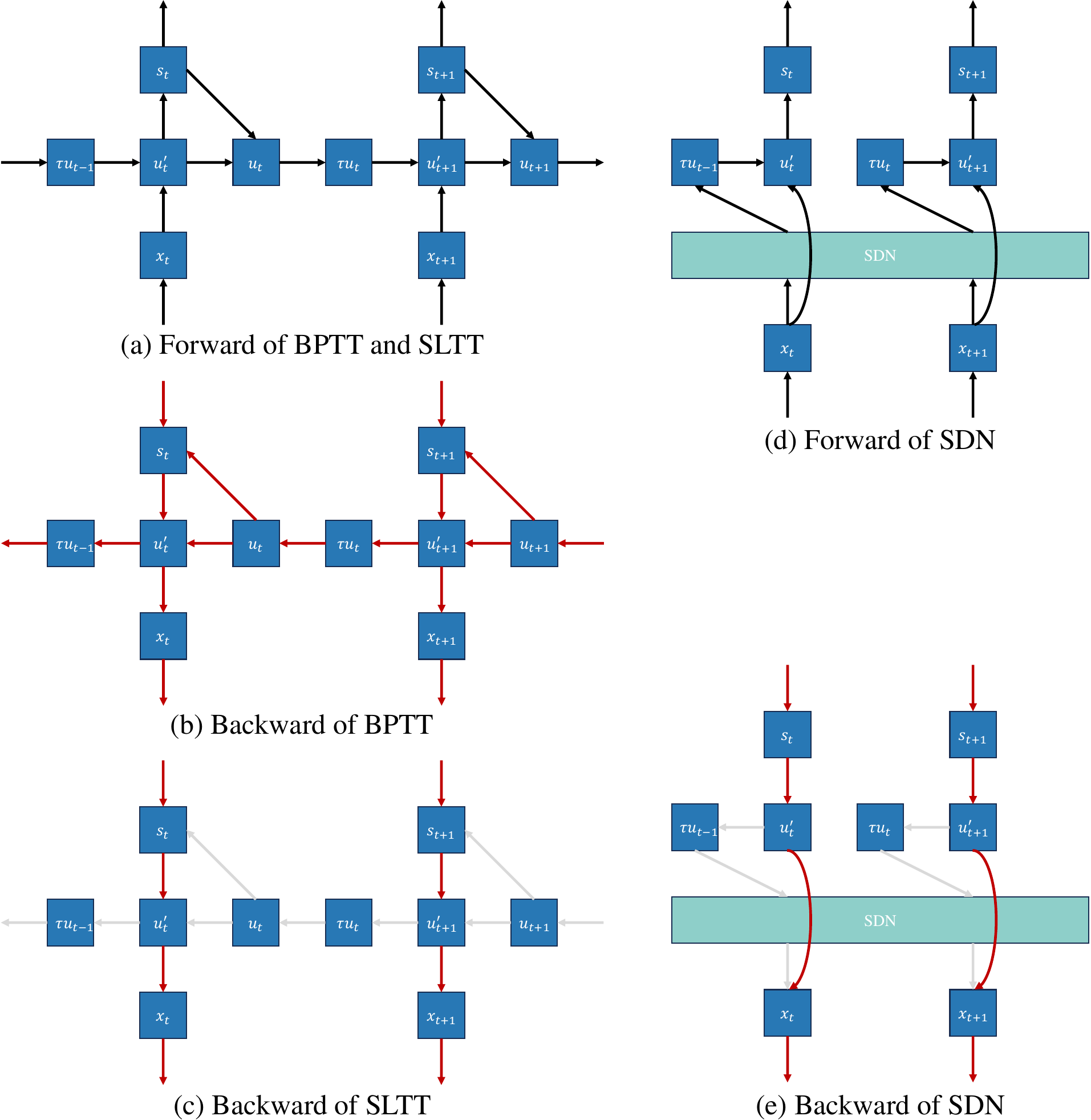}
    \caption{The forward and backward computational graphs of three training methods. The black lines in forward graphs and the red lines in backward graphs denote data flows;
        the gray lines are not data flows, although they have corresponding parts in the forward graphs.}
    \label{fig:backward}
\end{figure*}
We have mentioned three training methods in the main content: back-propagation through time (BPTT), spatial learning through time (SLTT) \cite{SLTT}, and parallel training with surrogate dynamic network (SDN). In this section, we derive the gradients of parameters under these training methods. For a clear explanation, we provide their forward and backward computational graphs in Fig. \ref{fig:backward}. Note that the forward graph of BPTT and SLTT have the same form (Fig. \ref{fig:backward}a), and the backward graph of SLTT and SDN have the same form (Fig. \ref{fig:backward}c \& e).

\subsection{Forward}
The LIF neuron with hard reset mechanism can be formulated as:
\begin{equation}
    u_t' = \tau u_{t - 1} + x_t
    \label{eq:bptt:li}
\end{equation}
\begin{equation}
    s_t = H(u_t' - v_{\mathrm{th}})
    \label{eq:bptt-spike}
\end{equation}
\begin{equation}
    u_t = u_t'(1 - s_t) + s_t u_r
    \label{eq:bptt-reset}
\end{equation}
where the $t$ denotes the step, $x$ denotes the input current, $u$ denotes the membrane potential, $s$ denotes the spike value (0 for non-spike and 1 for spike), and $v_{\mathrm{th}}$ denotes the threshold.
$u_r$ is the reset membrane potential and we set $u_r = 0$ in our experiments.
$H$ denotes the Heaviside function and we apply a surrogate gradient for its backward computation, denoted as $\dot{g}$.
For simplification, we denote $L$ as the loss, $\delta_t$ as the gradient of $s_t$, and this term is the only gradient from the next layer, not including the gradient from $u_t$.
Note that we ignore the layer index, as for each layer, if the gradient of $s_t$ is given, the other parameters have the same form of gradient in all layers.
Since we train our models with learnable $v_{\mathrm{th}}$, we also derive its gradient.

\subsection{Backward of BPTT}
For BPTT, the backward computational graph follows the form as the forward with inverse data flow, as depicted in Fig. \ref{fig:backward}b. We can get the gradients of the input and threshold by the following derivation:

\begin{align}
    \frac{\partial L}{\partial x_t}
    & = \frac{\partial L}{\partial u_t'} \frac{\partial u_t' }{\partial x_t} = \frac{\partial L}{\partial u_t'}  \\
    & =\frac{\partial L}{\partial s_t} \frac{\partial s_t }{\partial u_t'} + \frac{\partial L}{\partial u_t} \frac{\partial u_t }{\partial u_t'} + \frac{\partial L}{\partial u_t} \frac{\partial u_t }{\partial s_t} \frac{\partial s_t }{\partial u_t'} \\
    & = \delta_t \dot{g} + \frac{\partial L}{\partial u_t}[1 - s_t + (u_r - u'_t) \dot{g}]\\
    \label{eq:bptt}
    & = \delta_t \dot{g} + \tau \frac{\partial L}{\partial u_{t + 1}'}[1 - s_t + (u_r - u'_t) \dot{g}] \\
    \frac{\partial L}{\partial v_{\mathrm{th}}} 
    & = \sum_{t = 1}^{T}[\frac{\partial L}{\partial s_t} \frac{\partial s_t }{\partial v_{\mathrm{th}}} + \frac{\partial L}{\partial u_t} \frac{\partial u_t }{\partial s_t} \frac{\partial s_t }{\partial v_{\mathrm{th}}}]
    \\
    & = \sum_{t = 1}^{T}[\delta_t + \frac{\partial L}{\partial u_t}(u_r - u_t')]  (-\dot g)
\end{align}

In some cases, in order to accelerate the gradient calculation, $s_t$ is 'detached' from the gradient computation of $u_t$, which means the gradient of $s_t$ will not flow through $u_t$. In this case, the derivation can be expressed as:

\begin{align}
    \frac{\partial L}{\partial x_t} = \frac{\partial L}{\partial u_t'}
     & = \delta_t \dot{g} + \frac{\partial L}{\partial u_t}(1 - s_t)              \\
     & = \delta_t \dot{g} + \tau \frac{\partial L}{\partial u'_{t + 1}} (1 - s_t) \\
    \frac{\partial L}{\partial v_{\mathrm{th}}}
    &= \sum_{t = 1}^T\delta_t (-\dot g)
\end{align}

\subsection{Backward of SLTT}
Motivated by the gradient calculation of BPTT, SLTT further divides the gradient flow into a spatial and a temporal component. In equation \ref{eq:bptt}, the term $\delta_t \dot{g}$ is treated as the spatial component, as it is from the next layer. Similarly, the remaining part $\tau \frac{\partial L}{\partial u_{t + 1}'}[1 - s_t + (u_r - u'_t) \dot{g}]$ is treated as the temporal component, as it is from the next time step.
Note the term $\tau$ in the temporal component, which usually adopts a very small value, such that the temporal component is small compared to the spatial component. Therefore, in SLTT it was proposed that the temporal component can be omitted in the computational graph. In this case, the derivation of gradients become:
\begin{align}
    \frac{\partial L}{\partial x_t} 
    &= \frac{\partial L}{\partial u_t'}
    = \delta_t \dot{g}
    \label{eq:sltt-u} \\
    \frac{\partial L}{\partial v_{\mathrm{th}}} 
    &= \sum_{t = 1}^T\delta_t(-\dot g)
    \label{eq:sltt-vth}
\end{align}

According to this form, the path from $u_{t + 1}'$ to $u_t'$ is 'detached' from the computational graph.
Therefore, the whole computational graph for step range from 1 to $T$ is not necessary, and only the single-step computational graph needs to be retained, which reduces the memory of the training of SNN.

\subsection{Backward of SDN}
In our experiment, SDN is trained to predict the term $\tau u_{t - 1}$ in equation \ref{eq:bptt:li} when given the input current $x_{1:t}$, and it is switched to inference mode without backward gradient computation during the training of the SpikingSSM. Since SDN is designed for parallel computing, we formulate the equations with all variables treated as sequences, e.g., $x_{1:T}$ is the input current of all steps. In this case, the forward computation can be expressed as:
\begin{align}
    \tau u_{0:T-1} &=  f(x_{1:T})
    \label{eq:sdn-inference} \\
    u_{1:T}' &= \tau u_{0:T-1} + x_{1:T}
    \label{eq:sdn-li} \\
    s_{1:T} &= H(u_{1:T}' - v_{\mathrm{th}})\label{eq:sdn-fire} \\
\end{align}
where $f$ denotes SDN. Note that equation \ref{eq:bptt-reset} is not used here, as SDN directly predicts the membrane potential after the reset. Fig. \ref{fig:backward}d illustrates the corresponding forward computational graph.
For the backward graph, since there is no gradient computation in SDN, we can easily derive the gradients as the following:
\begin{equation}
    \frac{\partial L}{\partial x_{1:T}} = \frac{\partial L}{\partial u_{1:T}'}
    = \delta_{1:T} \dot{g}
    \label{eq:sdn-u}
\end{equation}

\begin{equation}
    \frac{\partial L}{\partial v_{\mathrm{th}}} = \sum_{t = 1}^T\delta_t(-\dot g)
    \label{eq:sdn-vth}
\end{equation}
These formulas are the same as those in SLTT.

\section{Architecture of SDN} \label{appendix:arch}
\begin{figure*}[ht]
\centering
	\subfloat[The architecture of SDN in the training mode.]{\includegraphics[width = 0.4\textwidth]{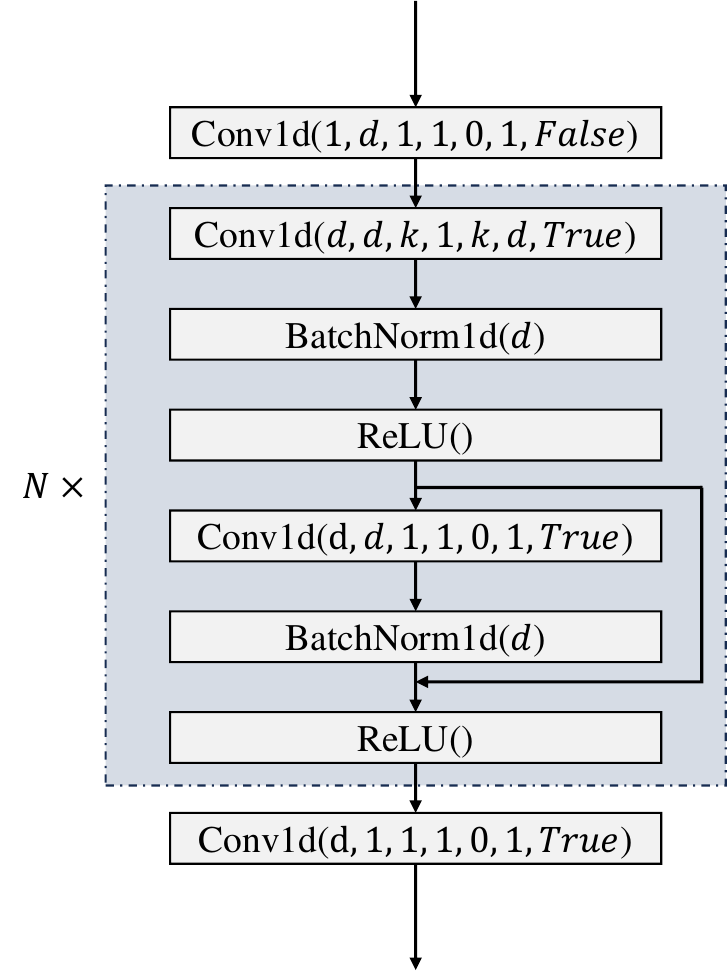}}
	\hfill
	\subfloat[The architecture of SDN in the inference mode.]{\includegraphics[width = 0.4\textwidth]{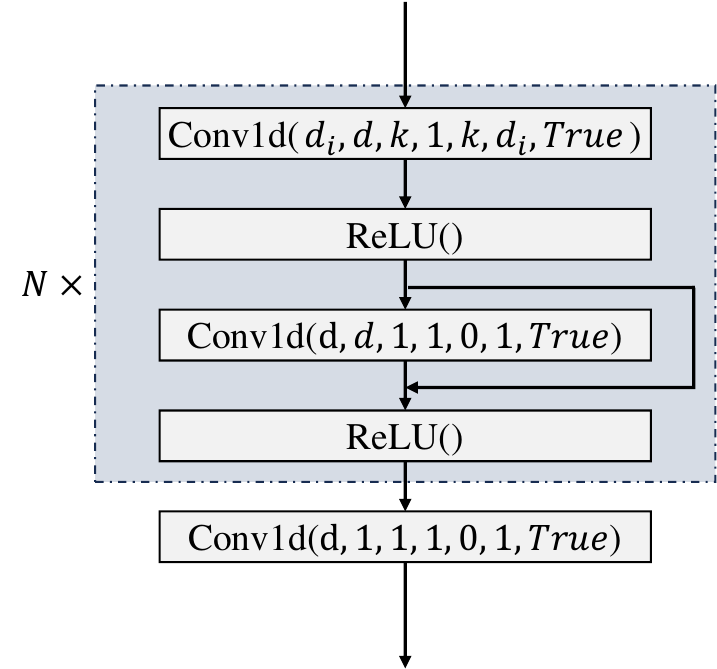}}
\caption{The architectures of SDN.}
\label{fig:sdn-arch}
\end{figure*}

The architecture of SDN during training and inference are shown in Fig. \ref{fig:sdn-arch}. For the SDN in training, the architecture has an encoder and a decoder at the beginning and the ending, respectively.
The 1-D convolution with kernel size of 1 can be viewed as a linear projection in the feature dimension. The numbers and characters of the 'Conv1d' in the figure corresponds to input channel, output channel, kernel size, stride, padding, groups and bias, respectively.
The value of hyper-parameters in these modules in our experiments are $N = 1$, $d = 8$, $k = 8$, $p = 8$ and $d_i = 1$.
Note that if $N > 1$, the hyper-parameter values in the first block keep unchanged but some values are changed in the rest blocks, i.e., $p = k - 1$ and $d_i = d$. After each group convolution, there is a slice operation to keep the first $L$ steps. In inference mode, the 'BatchNorm1d' layer is removed because of the fusion with the previous 'Conv1d' layer. The encoder is also removed because of the fusion with the first group convolution.

\section{Learnable Threshold for SDN} \label{appendix:vth}
In the main content, we claim that for SDN, the threshold can be trained equivalently by training a scaling factor of the inputs, we provide here the corresponding derivation.
Unroll equation \ref{eq:bptt:li} and equation \ref{eq:bptt-reset}, we can get:
\begin{align}
    u_t'
     & = \tau u_{t - 1}' (1 - s_{t - 1}) + x_t                                      \\
     & = \sum_{i = 1}^{t}\left[\tau^{t - i}x_i\prod_{j = i}^{t - 1}(1 - s_j)\right]
\end{align}
From the spiking condition, we have:
\begin{equation}
    u_t' = \sum_{i = 1}^{t}\left[\tau^{t - i}x_i\prod_{j = i}^{t - 1}(1 - s_j)\right] > v_{\mathrm{th}}
    \label{eq:unroll-spike}
\end{equation}
If $\alpha$ is a positive number, we can multiply it to both sides of the inequality and it still holds:
\begin{equation}
    \sum_{i = 1}^{t}\left[\tau^{t - i}(\alpha x_i)\prod_{j = i}^{t - 1}(1 - s_j)\right] > \alpha v_{\mathrm{th}}
\end{equation}
Therefore, if we set $\alpha$ to $\frac{1}{v_{\mathrm{th}}}$, we can get a fixed threshold 1 for SDN, which reduces the difficulty of network design.
For trainable threshold, we can first train SDN under $v_{\mathrm{th}} = 1$, and during the training of the task network assisted by SDN, we then divide the input currents by $v_{\mathrm{th}}$, thus training the scaling factor of the inputs.
In this case, the gradient of $v_{\mathrm{th}}$ change from equation \ref{eq:sdn-vth} to:
\begin{equation}
    \frac{\partial L}{\partial v_{\mathrm{th}}} = \sum_{t = 1}^T\delta_t \dot g (-\frac{x_t}{v_{\mathrm{th}}^2})
\end{equation}

\section{Other Experiments Results} \label{appendix:more}
In this section, we present additional results that were not included in the main content.

\subsection{DVS128 Gesture}
The DVS128 Gesture dataset, a neuromorphic dataset, consists of 11 hand gesture categories from 29 subjects under three illumination conditions. These images are based on event-driven recording, and we refer to previous work \cite{evnn2023} to preprocess the data format from event streams to frames. The results are reported in Table \ref{tb:DVS128-Gesture}.

\begin{table}[h]
    \centering
    \begin{tabular}{l c c}
        \hline
        \textbf{Model}            & \textbf{\#Params} & \textbf{Accuracy (\%)} \\ \hline
        EGRU\cite{evnn2023}       & 4.8 M     & 97.8                   \\
        Event-SSM\cite{Event-SSM} & 5.0 M     & 97.7                   \\
        SpikingSSM                & 3.0 M     & 97.1                   \\
        \hline
    \end{tabular}
    \caption{Performance comparison of SpikingSSM and other works on the DVS128 Gesture datasets.}
    \label{tb:DVS128-Gesture}
\end{table}

\subsection{Speech Commands}
The Speech Commands dataset \cite{sc} is a 35-class dataset of spoken English words. We follow the methodology of previous work \cite{gu2021efficiently} and report our results in Table \ref{tb:sc}.

\begin{table}[h]
    \centering
    \begin{tabular}{l c c}
        \hline
        \textbf{Model} & \textbf{\#Params} & \textbf{Accuracy (\%)} \\ \hline
        S4D-Lin        & 306K              & 96.25                  \\
        SpikingSSM     & 305K              & 96.09                  \\
        \hline
    \end{tabular}
    \caption{Performance comparison of SpikingSSM and other works on the Speech Commands datasets.}
    \label{tb:sc}
\end{table}

\subsection{Generalization of SDN}
We evaluate the generalization capabilities of SDN across datasets with varying parameters. Initially, we train the SDN on a dataset consisting of 50,000 samples drawn from $N(0, 1)$ with a length of 1024 for 100 epochs. The labels are generated using an LIF neuron with an attenuation coefficient of $\tau = 0.2$. Subsequently, the SDN is frozen (inference mode, with all parameters frozen) and evaluated on different datasets that vary in length, input distribution, and $\tau$ values. The results are summarized in Table \ref{tb:length}, Table \ref{tb:distribution}, and Table \ref{tb:tau}, respectively.
 
These tables demonstrate that the SDN exhibits robust performance across different input lengths and distributions. However, its performance degrades when faced with varying attenuation coefficients $\tau$. The latter phenomenon is understandable due to the change in temporal dynamics and could be mitigated if the SDN were trained the same $\tau$ value.

\begin{table}[h]
    \centering
    \begin{tabular}{c c c}
        \hline
        \textbf{Length} & \textbf{Accuracy (\%)} & \textbf{Loss} \\\hline
        1024            & 99.94966               & 0.000036      \\
        2048            & 99.94952               & 0.000036      \\
        4096            & 99.94943               & 0.000036      \\
        8192            & 99.94929               & 0.000036      \\
        16384           & 99.94921               & 0.000036      \\
        \hline
    \end{tabular}
    \caption{Test of SDN on datasets with different lengths.}
    \label{tb:length}
\end{table}

\begin{table}[h]
    \centering
    \begin{tabular}{c c c}
        \hline
        \textbf{Distribution} & \textbf{Accuracy (\%)} & \textbf{Loss} \\\hline
        $N(0, 1^2)$           & 99.95                  & 0.000036      \\
        $N(0, 2^2)$           & 99.93                  & 0.000071      \\
        $N(0, 3^2)$           & 99.91                  & 0.000247      \\
        $N(-1, 1^2)$          & 99.99                  & 0.000004      \\
        $N(-1, 2^2)$          & 99.97                  & 0.000027      \\
        $N(-1, 3^2)$          & 99.95                  & 0.000108      \\
        $N(1, 1^2)$           & 99.80                  & 0.000130      \\
        $N(1, 2^2)$           & 99.86                  & 0.000208      \\
        $N(1, 3^2)$           & 99.85                  & 0.000601      \\
        \hline
    \end{tabular}
    \caption{Test of SDN on datasets with different distributions.}
    \label{tb:distribution}
\end{table}

\begin{table}[h]
    \centering
    \begin{tabular}{c c c}
        \hline
        \textbf{$\tau$} & \textbf{Accuracy (\%)} & \textbf{Loss} \\\hline
        0.2             & 99.95                  & 0.000036      \\
        0.3             & 98.65                  & 0.004588      \\
        0.4             & 97.32                  & 0.020054      \\
        0.5             & 95.95                  & 0.052128      \\
        0.6             & 94.55                  & 0.111047      \\
        0.7             & 93.11                  & 0.214732      \\
        0.8             & 91.58                  & 0.406451      \\
        0.9             & 89.76                  & 0.870313      \\
        1.0             & 85.92                  & 15.63004      \\
        \hline
    \end{tabular}
    \caption{Test of SDN on datasets with different $\tau$.}
    \label{tb:tau}
\end{table}

\subsection{Breakdown of the Latency}
In the main content, we benchmark the training speed of different methods in Table \ref{tb:benchmark}. To analyze the reasons for the low latency of SDN, we conduct a breakdown experiment using a 4-layer model (encoder-4×(ssm-lif-linear)-decoder-norm) with three training methods on a 1K sequence.
The systematic breakdown of latency during the training process is presented in Table \ref{tb:latency}. Due to precision and order-of-magnitude issues in measuring latency, the latency of BPTT and SLTT is filled with 3 zeros after the decimal point for better comparison with other latency. Note that CUDA operations are executed asynchronously; therefore, we cannot directly sum CPU time and GPU time, nor can we consider either CPU time or GPU time as the program execution time.

The results demonstrate that the latency of SpikingSSMs is primarily determined by the temporal dynamics of spiking neurons. SDN reduces this latency to a level comparable to that of other modules. Furthermore, with the assistance of SDN, the membrane potential at all time steps can be accessed simultaneously, allowing the decision to fire or not to be determined in parallel. This is the reason why the latency of the 'fire' process in SDN ($0.095 ms$) is lower than that of BPTT ($83 ms$) and SLTT ($82 ms$). Note that the LI process is implemented as a for-loop in BPTT and SLTT, but in SDN, it is a parallelized CNN model. Therefore, with longer time steps, SDN will exhibit a more significant delay reduction effect.

\begin{table*}[h]
    \centering
    \begin{tabular}{lrrrr}
        \hline
        \textbf{Module}    & \textbf{CPU time (ms)} & \textbf{Avg CPU time (ms)} & \textbf{CUDA time (ms)} & \textbf{Avg CUDA time (ms)} \\\hline
        Encoder           & 1.978                  & 1.978                      & 0.023                   & 0.023                       \\
        SSM               & 5.042                  & 1.261                      & 0.849                   & 0.212                       \\
        \hline
        BPTT (LI)         & 694.000                & 173.000                    & 85.000                  & 21.000                      \\
        SLTT (LI)         & 678.000                & 169.000                    & 88.000                  & 22.000                      \\
        \textbf{SDN (LI)} & 3.307                  & 0.827                      & 1.499                   & 0.375                       \\
        \hline
        BPTT (F)          & 332.000                & 83.000                     & 40.000                  & 10.000                      \\
        SLTT (F)          & 330.000                & 82.000                     & 41.000                  & 10.000                      \\
        \textbf{SDN (F)}  & 0.379                  & 0.095                      & 0.056                   & 0.014                       \\
        \hline
        Output Linear     & 1.845                  & 0.923                      & 0.190                   & 0.095                       \\
        Norm              & 0.557                  & 0.139                      & 0.134                   & 0.034                       \\
        Decoder           & 0.213                  & 0.213                      & 0.029                   & 0.029                       \\
        \hline
    \end{tabular}
    \caption{The breakdown of the latency during the training process of a 4-layer model using different training methods is as follows. The 'LI' in parentheses represents the leaky integrate process, specifically the equations \ref{eq:bptt:li} and \ref{eq:bptt-reset} for BPTT and SLTT, and the equations \ref{eq:sdn-inference} and \ref{eq:sdn-li} for SDN. The 'F' in parentheses represents the fire process, specifically the equation \ref{eq:bptt-spike} for BPTT and SLTT, and the equation \ref{eq:sdn-fire} for SDN. Note that the latency of BPTT and SLTT is filled with 3 zeros after the decimal point for better comparison with other latency.}
    \label{tb:latency}
\end{table*}

\section{Membrane Potential Distribution Analysis}\label{appendix:mem-dis}
\begin{figure*}[h!]
    \centering
    \includegraphics[width=1.0\linewidth]{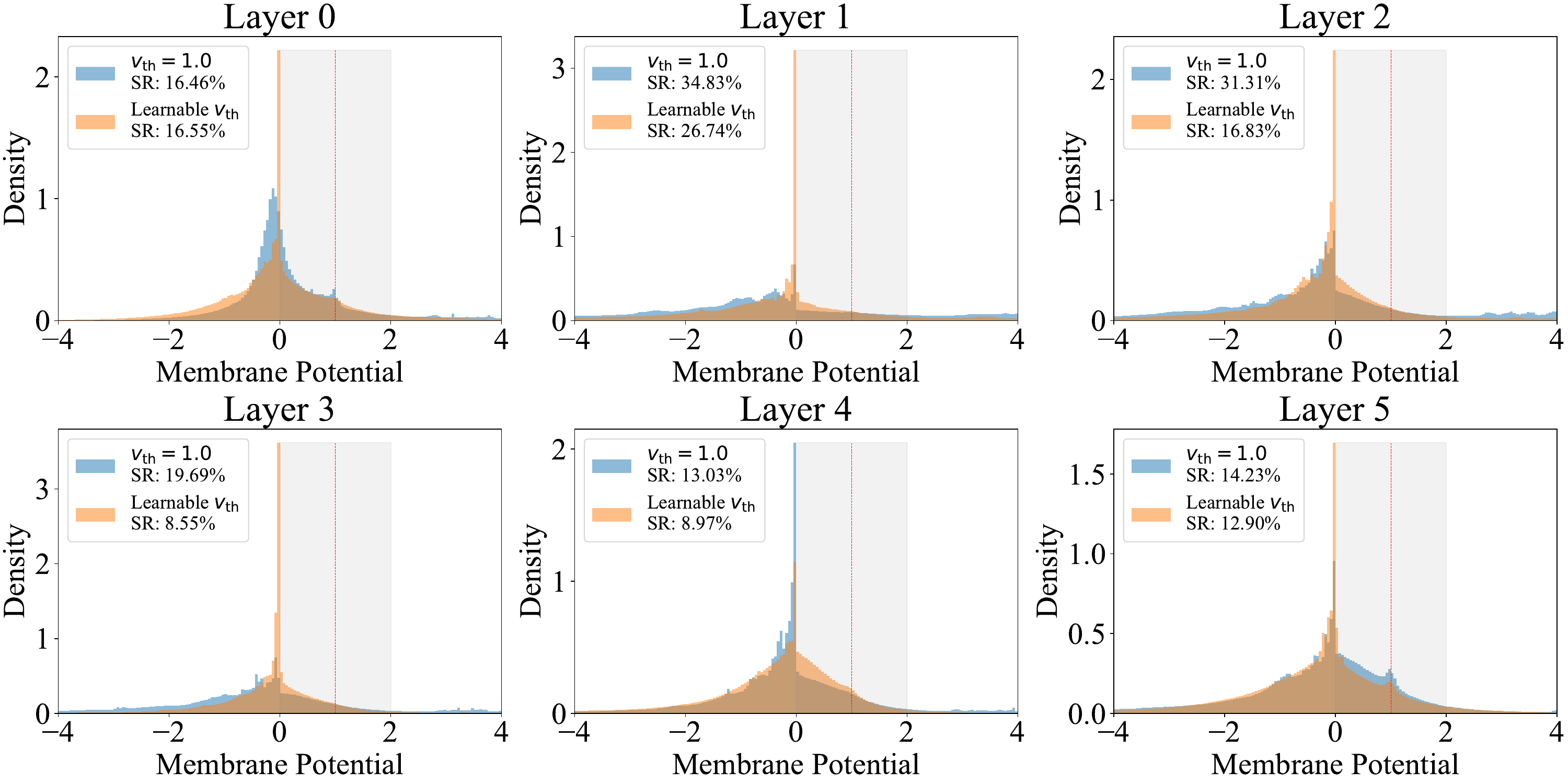}
    \caption{Membrane potential distributions, SR denotes spiking rate. The red line denotes the (effective) threshold. The shaded areas in the histograms represent the range where the surrogate activation function's gradient is nonzero.}
    \label{mem-potential}
\end{figure*}
In the LRA IMAGE task (sequential CIFAR10 task), we plot the membrane potential distributions of each layer of the SpikingSSM under both trainable and fixed thresholds ($v_{\mathrm{th}}$ = 1), as shown in Figure \ref{mem-potential}. The shaded areas in the histograms represent the range where the surrogate activation function's gradient is nonzero (for details of the SG function see the experimental settings section). For trainable thresholds, we shift the membrane potential of neurons with different thresholds correspondingly to ensure consistency in the non-zero gradient regions. Using trainable thresholds leads to marked decrease in spiking rate, which can be observed in most layers (Layer 1, 2, 3, 4, 5) where fixed membrane potential have more values above the thresholds. Remarkably, it can be also observed that the trainable threshold
facilitates a bimodal distribution of the membrane potential, i.e. membrane potentials are more gathered near '0' and '1', which reduces quantization error of spiking and improves information transmission of the SNN, consistent with previous findings \cite{guo2022recdis}.

For Layer 0 and 5, it can be observed that trainable thresholds leads to less gradient mismatch problem according to \cite{guo2022recdis}, i.e. fewer values within the non-zero SG gradient range. Moreover, the saturation problem \cite{guo2022recdis} is also better addressed, i.e. membrane potentials are less distributed outside the non-zero SG gradient range in corresponding layers (Layer 1, 2, 3, 4). The reduction in saturation phenomena supports the superiority of trainable thresholds over fixed thresholds in terms of sparsity and accuracy in the IMAGE task.

\section{Experimental Settings}\label{appendix:setting}
We list the specific parameter settings for all experiments in Table \ref{tab:lra-hyperparameters}, covering six tasks from the LRA, as well as sMNIST, psMNIST, and WikiText-103. The accuracy for all experiments is derived from the models that performed best on the validation set when evaluated on the test set. The SpikingSSM model utilizes the piecewise quadratic surrogate spiking function. The gradient is defined by:
\begin{equation}
    g'(x) =
    \begin{cases}
        0,                      & \text{if } |x| > \frac{1}{\alpha}    \\
        -\alpha^2 |x| + \alpha, & \text{if } |x| \leq \frac{1}{\alpha}
    \end{cases}
\end{equation}
and the primitive function is defined by:
\begin{equation}
    g(x) =
    \begin{cases}
        0,                                                   & \text{if } x < -\frac{1}{\alpha}     \\
        -\frac{1}{2} \alpha^2 |x|x + \alpha x + \frac{1}{2}, & \text{if } |x| \leq \frac{1}{\alpha} \\
        1,                                                   & \text{if } x > \frac{1}{\alpha}
    \end{cases}
\end{equation}
for our experiments, we set the value of $\alpha$ at 1.

\begin{table*}[ht]
    \centering
    \begin{tabular}{lccccccccccc}
        \hline
        \textbf{Dataset}    & \textbf{Depth} & \textbf{H} & \textbf{N} & \textbf{Norm} & \textbf{pNorm} & \textbf{Dropout} & \textbf{LR} & \textbf{BS} & \textbf{Epochs} & \textbf{WD} & \textbf{($\Delta_{min}$, $\Delta_{max}$)} \\
        \hline
        \textbf{ListOps}    & 8              & 128        & 64         & BN            & False          & 0                & 0.01        & 50          & 40              & 0.05        & (0.001, 0.1)                              \\
        \textbf{Text}       & 6              & 256        & 64         & BN            & True           & 0                & 0.01        & 16          & 32              & 0.01        & (0.001, 0.1)                              \\
        \textbf{Retrieval}  & 6              & 256        & 64         & BN            & True           & 0                & 0.01        & 64          & 20              & 0.01        & (0.001, 0.1)                              \\
        \textbf{Image}      & 6              & 512        & 64         & LN            & False          & 0.1              & 0.01        & 50          & 200             & 0.01        & (0.001, 0.1)                              \\
        \textbf{Pathfinder} & 6              & 256        & 64         & BN            & True           & 0                & 0.004       & 64          & 200             & 0.01        & (0.001, 0.1)                              \\
        \textbf{Path-X}     & 6              & 256        & 64         & BN            & True           & 0                & 0.0005      & 32          & 50              & 0.01        & (0.0001, 0.01)                            \\
        \hline
        \textbf{WT-103}     & 16             & 1024       & 64         & LN            & True           & 0.1              & 0.0005      & 1           & 200             & 0.01        & (0.001,0.1)                               \\
        \textbf{psMNIST}    & 4              & 256        & 64         & LN            & False          & 0.1              & 0.01        & 50          & 100             & 0.01        & (0.001,0.1)                               \\
        \textbf{sMNIST}     & 2              & 400        & 64         & LN            & False          & 0.1              & 0.01        & 50          & 100             & 0.01        & (0.001,0.1)                               \\
        \hline
        \textbf{DVS128-Gesture}    & 6              & 512        & 16         & LN            & False          & 0.1               & 0.005        & 32          & 200              & 0.05        & (0.001, 0.1)                              \\
        \textbf{SC}    & 6              & 128        & 32         & BN            & True         & 0                & 0.01        & 16          & 40              & 0.01        & (0.001, 0.1)                              \\
        \hline
    \end{tabular}
    \caption{
        The hyper-parameters of our experiments on these datasets. H denotes the model dimension, N denotes the state dimension, LR denotes learning rate, WD denotes weight decay and BS denotes the batch size. BN and LN refer to Batch Normalization and Layer Normalization.
    }
    \label{tab:lra-hyperparameters}
\end{table*}

\section*{Acknowledgments}
This work is supported in part by the National Key Research and Development Program of China (2021YFF1200800), the National Natural Science Foundation of China (Grant No. 62276121, 12326604) and the Science and Technology Innovation 2030-Major Project (Brain Science and Brain-Like Intelligence Technology) under Grant 2022ZD0208700.

\bibliography{aaai25}

\end{document}